\documentclass{article}
\usepackage{spconf,amsmath,epsfig}
\usepackage{booktabs}
\usepackage{epstopdf}
\usepackage{caption}
\usepackage{bm}
 \usepackage{float}
 \usepackage{amsmath}
\usepackage{accents}
\usepackage{statex}
\usepackage{ragged2e}
\usepackage{cite}
\usepackage[numbers,sort&compress]{natbib}
\usepackage{amssymb}
\usepackage{multirow}
\usepackage{booktabs}
\newcommand{\argmin}{\mathop{\mathrm{arg\,min}}}

\title{Constrained low-tubal-rank tensor recovery for hyperspectral images mixed noise removal by
bilateral random projections}
\name{Hao Zhang$^{\dagger}$,
Xi-Le Zhao$^{\dagger\ast}$,
Tai-Xiang Jiang$^{\dagger}$,
Michael Kwok-Po Ng$^{\ddagger}$\thanks{The research is supported by NSFC (61876203, 61772003),
the HKRGC GRF (1202715, 12306616,
12200317), and HKBU RC-ICRS/16-17/03.}\thanks{\hspace{-0.5cm}$^{\ast}$Corresponding author: xlzhao122003@163.com.}}

\address{\normalsize $^{\dagger}$School of Mathematical Sciences/Research Center for Image and Vision Computing, \\ \normalsize University of Electronic Science and Technology of China, Chengdu, P. R. China\\
\normalsize $^{\ddagger}$Department of Mathematics, Hong Kong Baptist University, Kowloon Tong, Hong Kong}

\usepackage{amsfonts}
\begin{document}
%\ninept
\maketitle
\begin{abstract}
In this paper, we propose a novel low-tubal-rank tensor recovery model,
which directly constrains the tubal rank prior for effectively removing the mixed
Gaussian and sparse noise in hyperspectral images. The constraints of tubal-rank and sparsity can govern the solution
of the denoised tensor in the recovery procedure. To solve the constrained low-tubal-rank model, we
develop an iterative algorithm based on
bilateral random projections
to efficiently solve the proposed model. The advantage of random projections
is that the approximation of the low-tubal-rank tensor can be obtained quite accurately
in an inexpensive manner. Experimental examples for hyperspectral image denoising
are presented to demonstrate the effectiveness and efficiency of the proposed method.
\end{abstract}
\begin{keywords}
Tensor, bilateral random projections, low-tubal-rank, mixed noise, hyperspectral images.
\end{keywords}

\vspace{-0.1cm}
\section{INTRODUCTION}
\vspace{-0.2cm}
\label{sec:intro}
\noindent Hyperspectral remote sensing images are widely used in various applications \cite{Jiang1}. However, hyperspectral images (HSIs) in practice are often inevitably corrupted by several types of noises, such as Gaussian noise, sparse noise, stripes, and deadlines. Consequently, the applications are severely influenced. Therefore, it is essential to develop effective and efficient methods for HSIs denoising task like \cite{QY,YC}.

HSIs are spatially and spectrally correlated resulting in low-rankness. Many denoising methods are devoted to preserving the low-rank structure of the clean HSIs. One classical way is unfolding HSIs to matrices, such as low-rank matrix recovery (LRMR) \cite{LRMR}, nonconvex regularizer with weighted Schatten \emph{p}-norm (WSNLRMA) \cite{WSNLRMA}, and
low-rank subspace representation methods \cite{FHYDE,jiang2}. However, unfolding HSIs to matrices will destroy the
intrinsic structures.

Since tensor can better express more complex intrinsic structures of the higher-order data, the related researches have received considerable attention.
A corrupted HSI is a three-way tensor, which can be decomposed as a clean part, a sparse noise part, and a Gaussian noise part (See Fig.1). The first and second dimensions of HSIs are corresponding to
the spatial information while the third dimension reflects the spectral information. Spatial-spectral information can be simultaneously exploited by tensor-based methods. Based on
\begin{figure}[t]
  \begin{center}
  \includegraphics[width=0.48\textwidth]{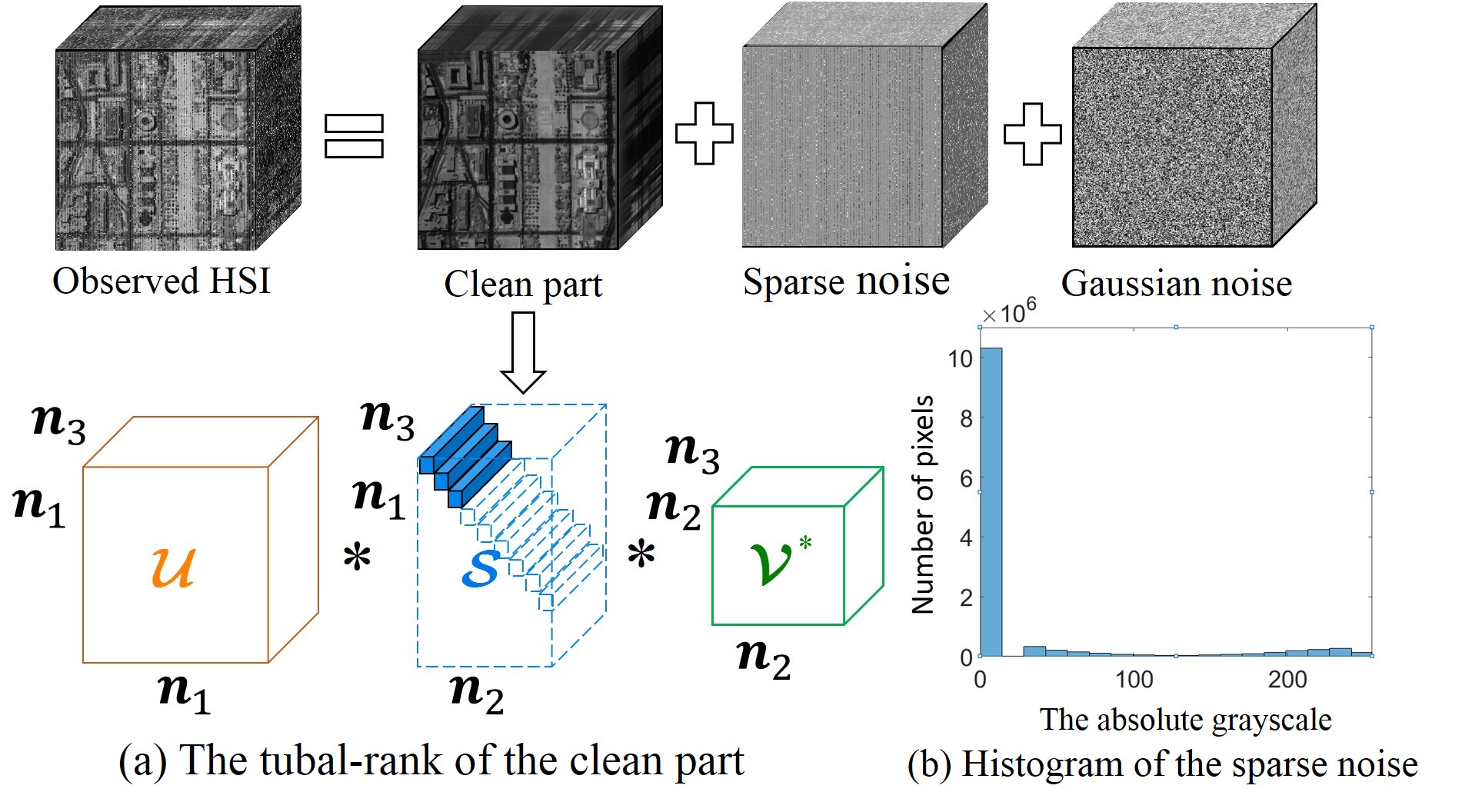}
  \end{center}
  \vspace{-0.7cm}
  \caption{Characteristics of the results recovered by CLTRTR. (a) Illustration of the number of non-zero tubes of the clean part, when the tubal rank prior is 3. (b) Illustration of the number of non-zero entries of the sparse noise part.}
 \label{motiv}
  \vspace{-0.5cm}
\end{figure}
different decomposition schemes, several low-rank tensor recovery methods have been proposed, such as low-$n$-rank tensor approximation (LRTA) \cite{LRTA} and the rank-1 tensor decomposition method \cite{R-RANK}.

The tensor tubal rank, based on the tensor singular value decomposition (t-SVD), is magnetic for well characterizing the inherent low-rank structure of a tensor.  Tensor nuclear norm (TNN) was considered formulating a low-rank tensor recovery model (LRTR) \cite{LRTR}  for effective HSIs denoising. Although TNN as a popular surrogate of the tubal rank has obtained promising results,  it is just a suboptimal surrogate of the tubal rank. For the large size HSIs, LRTR suffers from the heavy burden of computing singular value decomposition (SVD) within each frontal slice. These motivate us to derive a novel constrained  low-tubal-rank tensor recovery model with a fast and accurate algorithm for efficient HSIs mixed noise removal.

The main contributions of this work are two folds: (1) we propose a novel constrained low-tubal-rank tensor recovery model (CLTRTR) for effective HSI mixed Gaussian and sparse noise removal, which directly constrains the tubal rank of the target HSI (Fig.\ref{motiv} illustrates characteristics of the
results generated by CLTRTR)); (2) we design a tensor bilateral random projections algorithm (t-BRP) to efficiently solve the proposed model, which finely and quickly
approximates the low-tubal-rank tensor.

The rest of this paper is given as follows. Section \ref{sec:nota} introduces notations and preliminaries. Section \ref{sec:RES} gives the proposed model and algorithm. Section \ref{sec:exp} reports the results of numerical experiments. Section \ref{sec:concl} concludes this paper.

\vspace{-0.1cm}
\section{NOTATIONS AND PRELIMINARIES}
\vspace{-0.2cm}
\label{sec:nota}

\noindent For a three-way tensor $\bm{\mathcal{A}} \in \mathbb{R}$$^{n_{1} \times n_{2} \times n_{3}}$, the \emph{i}-th frontal slice is represented by $\emph{\textbf{A}}^{(\emph{i})}$. We denote $\bm{\mathcal{\bar{A}}}$  as the tensor generated by Discrete Fourier Transformation (DFT) on each tube of $\bm{\mathcal{A}}$. We denote $\bar{\emph{\textbf{A}}}$ as a block diagonal matrix whose \emph{i}-th block on the diagonal as the \emph{i}-th frontal slice $\bar{\emph{\textbf{A}}}^{(\emph{i})}$ of $\bm{\mathcal{\bar{A}}}$. Meanwhile, the block circulant matrix $\texttt{bcirc}$($\bm{\mathcal{A}}$)  is defined as
\begin{equation*}
\setlength{\abovedisplayskip}{5pt}
\setlength{\arraycolsep}{0.5pt}
\begin{aligned}
\texttt{bcirc}(\bm{\mathcal{A}})=
\begin{pmatrix}
\emph{\textbf{A}}^{(1)} &  \emph{\textbf{A}}^{(n_{3})}  & \cdots   & \emph{\textbf{A}}^{(2)}  \\
\emph{\textbf{A}}^{(2)} & \emph{\textbf{A}}^{(1)} & \cdots & \emph{\textbf{A}}^{(3)}  \\
\vdots  & \vdots  & \ddots & \vdots  \\
\emph{\textbf{A}}^{(n_{3})} & \emph{\textbf{A}}^{(n_{3}-1)}  & \cdots  & \emph{\textbf{A}}^{(1)} \\
\end{pmatrix}.
\end{aligned}
\setlength{\belowdisplayskip}{5pt}
\end{equation*}
The block circulant matrix can be block diagonalized, i.e.,
\begin{equation}
\setlength{\abovedisplayskip}{5pt}
(\emph{\textbf{F}}_{n_{3}} \otimes \emph{\textbf{I}}_{n_{1}}) \cdot \texttt{bcirc}(\bm{\mathcal{A}}) \cdot (\emph{\textbf{F}}^{-1}_{n_{3}} \otimes \emph{\textbf{I}}_{n_{2}}) = \bar{\emph{\textbf{A}}},
\label{Kro}
\setlength{\belowdisplayskip}{5pt}
\end{equation}
where $\otimes$ is the Kronecker product and $\emph{\textbf{F}}_{n_{3}}$ is the DFT matrix.\vspace{0.1cm}\\
$\textbf{Definition 2.1. (\emph{Tensor product})}$ \cite{def} For $\bm{\mathcal{A}} \in \mathbb{R}^{n_1 \times n_2 \times n_3}$ \emph{and} $\bm{\mathcal{B}} \in \mathbb{R}^{n_2 \times l \times n_3}$, the tensor product is defined to be a $n_1 \times l \times n_3$ tensor
\begin{equation}
\setlength{\abovedisplayskip}{5pt}
\bm{\mathcal{A}} \ast \bm{\mathcal{B}} = \texttt{fold}(\texttt{bcirc}(\bm{\mathcal{A}}) \cdot \texttt{unfold}(\bm{\mathcal{B}})).\label{tpro}
\setlength{\belowdisplayskip}{5pt}
\end{equation}
$\textbf{Definition 2.2. (\emph{Inverse of tensor})}$  \cite{def}  The inverse of a tensor  $\bm{\mathcal{A}} \in \mathbb{R}^{n \times n \times n_3}$  is written as $\bm{\mathcal{A}}^{-1}$ , satisfying $\bm{\mathcal{A}} ^{-1} \ast \bm{\mathcal{A}} = \bm{\mathcal{A}} \ast \bm{\mathcal{A}} ^{-1} = \bm{\mathcal{I}}.$ $\bm{\mathcal{I}}$ is the identity tensor whose first frontal slice is the $n \times  n$ identity matrix. \vspace{0.1cm}\\
$\textbf{Definition 2.3. (\emph{t-SVD})}$ \cite{def} $\bm{\mathcal{A}} \in \mathbb{R}$$^{n_{1} \times n_{2} \times n_{3}}$ can be factored as
\begin{equation}
\setlength{\abovedisplayskip}{5pt}
\bm{\mathcal{A}} = \bm{\mathcal{U}} \ast \bm{\mathcal{S}} \ast \bm{\mathcal{V}}^\ast,
\setlength{\belowdisplayskip}{5pt}
\end{equation}
where $\bm{\mathcal{U}} \in \mathbb{R}$$^{n_{1} \times n_{1} \times n_{3}}$ and $\bm{\mathcal{V}} \in \mathbb{R}$$^{n_{2} \times n_{2} \times n_{3}}$ are orthogonal,
$\bm{\mathcal{V}}^\ast$ is the conjugate transpose of $\bm{\mathcal{V}}$. $\bm{\mathcal{S}} \in \mathbb{R}$$^{n_{1} \times n_{2} \times n_{3}}$ is a $f$-diagonal tensor, whose each frontal slice is a diagonal matrix.\vspace{0.1cm}
\\$\textbf{Definition 2.3. (\emph{Tensor multi-rank and tubal rank})}$ \cite{def} The tensor multi-rank of $\bm{\mathcal{A}} \in \mathbb{R}$$^{n_{1} \times n_{2} \times n_{3}}$ is a vector $\emph{\textbf{r}} \in \mathbb{R}^{n_3}$, whose $i$-th element is the rank of the $i$-th frontal slice of $\bm{\mathcal{\bar{A}}}$, i.e., $\emph{r}_{\emph{i}} = \textrm{rank}(\bar{\emph{\textbf{A}}}^{(\emph{i})})$. The tensor tubal rank, denoted as $\mathrm{rank}_{\mathrm{t}}(\bm{\mathcal{A}})$, is defined as the number of non-zero tubes of $\bm{\mathcal{S}}$, where $\bm{\mathcal{S}}$ is from the t-SVD of $\bm{\mathcal{A}} = \bm{\mathcal{U}} \ast \bm{\mathcal{S}} \ast \bm{\mathcal{V}}^\ast$. That is
\begin{equation}
\setlength{\abovedisplayskip}{5pt}
\mathrm{rank}_{\mathrm{t}}(\bm{\mathcal{A}}) = \#\{i : \bm{\mathcal{S}}(i,i,:) \neq 0\} = \max_{i} r_i.
\setlength{\belowdisplayskip}{5pt}
\end{equation}
$\textbf{Definition 2.4. (\emph{TNN})}$ \cite{def} The TNN of a tensor $\bm{\mathcal{A}} \in \mathbb{R}^{n_1 \times n_2 \times n_3}$, denoted as $\rVert\bm{\mathcal{A}}\rVert_{\textrm{TNN}}$, is defined as the sum of singular values of all the frontal slices of $\bm{\mathcal{\bar{A}}}$, i.e.,
\begin{equation}
\setlength{\abovedisplayskip}{5pt}
\rVert\bm{\mathcal{A}}\rVert_{\textrm{TNN}} = \sum\nolimits_{i = 1}^{n_3} \rVert\bar{\emph{\textbf{A}}}^{(\emph{i})}\rVert_\ast.
\setlength{\belowdisplayskip}{5pt}
\end{equation}

It can been seen that the tubal rank is equal to the largest element of the multi-rank and TNN approximates the $l_1$-norm of the tensor multi-rank. Although TNN simplifies the recovery problem, it causes unavoidable bias.\vspace{0.1cm}
\\$\textbf{Remark 2.1}$ For $\bm{\mathcal{A}} \in \mathbb{R}^{n_{1} \times n_{2} \times n_{3}}$, $\mathrm{rank}_{\mathrm{t}}(\bm{\mathcal{A}}) \leq \textrm{min}(n_1, n_2)$ and $\mathrm{rank}_{\mathrm{t}}(\bm{\mathcal{A}} \ast \bm{\mathcal{B}}) \leq \textrm{min}(\mathrm{rank}_{\mathrm{t}}(\bm{\mathcal{A}}), \mathrm{rank}_{\mathrm{t}}(\bm{\mathcal{B}}))$.

\vspace{-0.1cm}
\section{MAIN RESULTS}
\vspace{-0.2cm}
\label{sec:RES}

\subsection{The proposed model}
\vspace{-0.1cm}
Let $\bm{\mathcal{X}} \in \mathbb{R}^{n_1 \times n_2 \times n_3}$ denotes an observed HSI, which can be expressed as the sum of three parts, i.e.,
\begin{equation}
\setlength{\abovedisplayskip}{5pt}
\bm{\mathcal{X}} = \bm{\mathcal{L}} + \bm{\mathcal{S}} + \bm{\mathcal{N}},
\setlength{\belowdisplayskip}{5pt}
\end{equation}
where $\bm{\mathcal{L}}$ is the clean HSI, $\bm{\mathcal{S}}$ is the sparse noise, and $\bm{\mathcal{N}}$ is the Gaussian noise. The goal of HSI denoising is to recover $\bm{\mathcal{L}}$ from observed $\bm{\mathcal{X}}$. Assuming that $\bm{\mathcal{L}}$ is low-tubal-rank with $\mathrm{rank}_\mathrm{{t}}(\bm{\mathcal{L}}) \leq r$ and the number of non-zero elements of $\bm{\mathcal{S}}$ is no more than $k$, i.e., $\mathrm{card}(\bm{\mathcal{S}}) \leq k$,
we formulate the model
\begin{equation}
\setlength{\abovedisplayskip}{5pt}
\begin{aligned}
\min_{\bm{\mathcal{L}}, \bm{\mathcal{S}}}\;\;&\rVert\bm{\mathcal{X}} - \bm{\mathcal{L}} - \bm{\mathcal{S}}\rVert_{F}^2,\\
\text{s.t.}\;\;&\mathrm{rank}_\mathrm{{t}}(\bm{\mathcal{L}}) \leq r, \mathrm{card}(\bm{\mathcal{S}}) \leq k.\\
\end{aligned}\label{my_model}
\setlength{\abovedisplayskip}{5pt}
\end{equation}

\vspace{-0.6cm}
\subsection{The proposed algorithm}
\vspace{-0.1cm}
Given $r$ and $k$, (\ref{my_model}) can be transformed into solving the following two subproblems alternately until convergence:
\begin{equation}
\setlength{\abovedisplayskip}{5pt}
\bm{\mathcal{L}}^{t} = \argmin_{\mathrm{rank}_\mathrm{{t}}(\bm{\mathcal{L}})\leq r}\rVert\bm{\mathcal{X}} - \bm{\mathcal{L}} - \bm{\mathcal{S}}^{t-1}\rVert_{F}^2,
\setlength{\belowdisplayskip}{5pt}\label{proL}
\end{equation}
\begin{equation}
\setlength{\abovedisplayskip}{5pt}
\bm{\mathcal{S}}^{t} = \argmin_{\mathrm{card}(\bm{\mathcal{S}})\leq k} \,\rVert\bm{\mathcal{X}} - \bm{\mathcal{L}}^{t} - \bm{\mathcal{S}}\rVert_{F}^2.
\setlength{\belowdisplayskip}{5pt}\label{proS}
\end{equation}
 $\bm{\mathcal{L}}^{t}$ can be obtained by t-SVD of $\bm{\mathcal{X}}-\bm{\mathcal{S}}^{t-1}$, due to the property that truncated t-SVD is optimal for
data approximation. However, SVDs within each slice are the main computation burden at each iteration. To efficiently solve (\ref{proL}), we design a t-BRP algorithm, which approximates the truncated t-SVD.  \vspace{0.1cm}  \\
$\textbf{Definition 3.1. (\emph{t-BRP})}$ For $\bm{\mathcal{X}} \in \mathbb{R}^{n_1 \times n_2 \times n_3}$ ($n_1 > n_2$), the t-BRP of $\bm{\mathcal{X}}$ can be constructed, i.e., $\bm{\mathcal{Y}}_1 = \bm{\mathcal{X}} \ast \bm{\mathcal{A}}_1 $ and $\bm{\mathcal{Y}}_2 = \bm{\mathcal{X}}^{\ast} \ast \bm{\mathcal{A}}_2$, wherein $\bm{\mathcal{A}}_1 \in \mathbb{R}^{n_2 \times r \times n_3}$ and $\bm{\mathcal{A}}_2 \in \mathbb{R}^{n_1 \times r \times n_3}$ are random tensors. \vspace{0.1cm}

The tubal rank-$r$ approximation of $\bm{\mathcal{X}}$ can be conducted as
\begin{equation}
\setlength{\abovedisplayskip}{5pt}
\bm{\mathcal{L}} = \bm{\mathcal{Y}}_1 \ast (\bm{\mathcal{A}}_2^{\ast} \ast \bm{\mathcal{Y}}_1 )^{-1} \ast \bm{\mathcal{Y}}_2^\ast,
\setlength{\belowdisplayskip}{5pt}\label{pro2L}
\end{equation}
where $\bm{\mathcal{A}}_2^{\ast}$ is the conjugate transpose of $\bm{\mathcal{A}}_2$ and (\ref{pro2L}) is a approximation of the truncated t-SVD as the explanation in \cite{cop}.

Recalling (\ref{Kro}) and (\ref{tpro}), we construct the matrix bilateral random projections of each $\bar{\emph{\textbf{X}}}^{(i)}$  as follows.
For $i = 1, \ldots , n_3$, let $\bar{\emph{{\textbf{Y}}}}_1^{(i)} = \bar{\emph{\textbf{X}}}^{(i)}\bar{\emph{\textbf{A}}}_1^{(i)}$, $\bar{\emph{{\textbf{Y}}}}_2^{(i)} = (\bar{\emph{\textbf{X}}}^{(i)})^\ast\bar{\emph{\textbf{A}}}_2^{(i)}$, in which $ \bar{\emph{\textbf{A}}}_1^{(i)} \in \mathbb{C}^{n_2 \times r}$ and $ \bar{\emph{\textbf{A}}}_2^{(i)} \in \mathbb{C}^{n_1 \times r}$ are random matrices. The low-rank approximation of $\bar{\emph{\textbf{X}}}^{(i)}$ is obtained by $\bar{\emph{\textbf{L}}}^{(i)} = \bar{\emph{{\textbf{Y}}}}_1^{(i)}[(\bar{\emph{{\textbf{A}}}}_2^{(i)})^\ast\bar{\emph{{\textbf{Y}}}}_1^{(i)}]^{-1}\bar{\emph{{\textbf{Y}}}}_2^{(i)} $. $\bar{\emph{{\textbf{Y}}}}_1 = \bar{\emph{{\textbf{X}}}}\bar{\emph{{\textbf{A}}}}_1$, $\bar{\emph{{\textbf{Y}}}}_2 = \bar{\emph{{\textbf{X}}}}^\ast\bar{\emph{{\textbf{A}}}}_2$, and
\begin{equation}
\setlength{\abovedisplayskip}{5pt}
\bar{\emph{{\textbf{L}}}} = \bar{\emph{{\textbf{Y}}}}_1(\bar{\emph{{\textbf{A}}}}_2^{\ast}\bar{\emph{{\textbf{Y}}}}_1 )^{-1}\bar{\emph{{\textbf{Y}}}}_2^\ast.
\setlength{\belowdisplayskip}{5pt}
\end{equation}
Then we have that $(\emph{{\textbf{F}}}_{n_3}^{-1} \otimes \emph{{\textbf{I}}}_{n_1} )\bar{\emph{{\textbf{X}}}}(\emph{{\textbf{F}}}_{n_3} \otimes \emph{{\textbf{I}}}_{n_2} )$, $(\emph{{\textbf{F}}}_{n_3}^{-1} \otimes \emph{{\textbf{I}}}_{n_2} ) \bar{\emph{{\textbf{A}}}}_{1}(\emph{{\textbf{F}}}_{n_3} \otimes \emph{{\textbf{I}}}_{r} )$, $(\emph{{\textbf{F}}}_{n_3}^{-1} \otimes \emph{{\textbf{I}}}_{n_2} ) \bar{\emph{{\textbf{X}}}}^\ast(\emph{{\textbf{F}}}_{n_3} \otimes \emph{{\textbf{I}}}_{n_1} )$, $(\emph{{\textbf{F}}}_{n_3}^{-1} \otimes \emph{{\textbf{I}}}_{n_1} ) \bar{\emph{{\textbf{A}}}}_{2}(\emph{{\textbf{F}}}_{n_3} \otimes \emph{{\textbf{I}}}_{r} )$, $(\emph{{\textbf{F}}}_{n_3}^{-1} \otimes \emph{{\textbf{I}}}_{n_1} ) \bar{\emph{{\textbf{Y}}}}_1 (\emph{{\textbf{F}}}_{n_3} \otimes \emph{{\textbf{I}}}_{r} )$, $[(\emph{{\textbf{F}}}_{n_3}^{-1} \otimes \emph{{\textbf{I}}}_{r} )\bar{\emph{{\textbf{A}}}}_2^\ast (\emph{{\textbf{F}}}_{n_3} \otimes \emph{{\textbf{I}}}_{n_1} )(\emph{{\textbf{F}}}_{n_3}^{-1} \otimes \emph{{\textbf{I}}}_{n_1} )\bar{\emph{{\textbf{Y}}}}_1(\emph{{\textbf{F}}}_{n_3} \otimes \emph{{\textbf{I}}}_{r} )]^{-1}$, and $(\emph{{\textbf{F}}}_{n_3}^{-1} \otimes \emph{{\textbf{I}}}_{r} )\bar{\emph{{\textbf{Y}}}}_2^\ast(\emph{{\textbf{F}}}_{n_3} \otimes \emph{{\textbf{I}}}_{n_2} )$ are real block circulant matrices. We obtain expressions for $\texttt{bcirc}(\bm{\mathcal{Y}}_1)$, $\texttt{bcirc}(\bm{\mathcal{Y}}_2)$, and $\texttt{bcirc}(\bm{\mathcal{L}})$. Folding up these results, the $\bm{\mathcal{L}}$ can be given.

Considering a three-way tensor $\bm{\mathcal{X}} \in \mathbb{R}^{n_1 \times n_2 \times n_3}$ with $\mathrm{rank}_\mathrm{{t}}(\bm{\mathcal{L}}) \leq r$, the computational complexity of the developed t-BRP is $O(n_1n_2n_3\log n_3 + r^2(n_2 + r)n_3 + n_1n_2n_3r)$, while t-SVD is $O(n_1n_2n_3\log n_3 + \min(n_1n_2^2, n_1^2n_2)n_3)$. Referring to the Remark 2.1, the parameter $r$ imposes a direct tubal rank constraint upon the estimated low-rank tensor $\bm{\mathcal{L}}$.

As for the subproblem (\ref{proS}), $\bm{\mathcal{S}}^{t}$ is updated as
\begin{equation}
\setlength{\abovedisplayskip}{5pt}
\bm{\mathcal{S}}^{t} = \mathcal{H}_{k}(\bm{\mathcal{X}}-\bm{\mathcal{L}}^{t}),
\setlength{\belowdisplayskip}{5pt}
\end{equation}
wherein $\mathcal{H}_{k}(\bm{\mathcal{X}}-\bm{\mathcal{L}}^{t})$ denotes the entry-wise hard thresholding operator, which sets all but the largest $k$ elements of $|\bm{\mathcal{X}}-\bm{\mathcal{L}}^{t}|$ to zero.

The proposed algorithm is summarized in Algorithm 1. The objective value $\rVert\bm{\mathcal{X}} - \bm{\mathcal{L}} - \bm{\mathcal{S}}\rVert_{F}^2$ converges to a local minimum based on the framework in \cite{Godec}.
\begin{table}[h]
\renewcommand\arraystretch{0.95}\vspace{-0.1cm}
 \begin{tabular}{lcl}
  \toprule
  $\textbf{Algorithm 1}$: CLTRTR for HSIs denoising. \ \ \ \ \ \  \ \ \ \ \ \ \ \ \ \ \ \
 \\
  \midrule
\textbf{Input:} $\bm{\mathcal{X}}$, $r$, $k$, $\epsilon$ \\[-2pt]
\textbf{Output:} $\bm{\mathcal{L}}, \bm{\mathcal{S}}$ \ \ \ \ \ \ \ \  \  \ \ \ \ \\[-2pt]
\textbf{Initialize:} $\bm{\mathcal{L}}^{0} = \bm{\mathcal{S}}^{0} = 0, t = 0$ \ \ \ \ \ \ \ \  \  \ \ \ \ \\[-2pt]
\textbf{While} $\rVert\bm{\mathcal{X}} - \bm{\mathcal{L}}^{t} - \bm{\mathcal{S}}^{t}\rVert_{F}^{2}/\rVert\bm{\mathcal{X}}\rVert_{F}^{2} > \epsilon$ \textbf{do} \\[-2pt]
\ \ \  t = t+1;\\[-2pt]
\ \ \ $\bm{\mathcal{Y}}_1 = (\bm{\mathcal{X}} - \bm{\mathcal{S}}^{t-1}) \ast \bm{\mathcal{A}}_1$, $\bm{\mathcal{A}}_2 = \bm{\mathcal{Y}}_1$, \\[-2pt]
\ \ \ $\bm{\mathcal{Y}}_2 = (\bm{\mathcal{X}} - \bm{\mathcal{S}}^{t-1})^\ast \ast \bm{\mathcal{A}}_2$;\\[-2pt]
\ \ \ $\textbf{If}\  \mathrm{rank}_\mathrm{{t}}(\bm{\mathcal{A}}_2^\ast \ast \bm{\mathcal{Y}}_2) < r$,\\[-2pt]
\ \ \ $\textbf{then}\ r = \mathrm{rank}_\mathrm{{t}}(\bm{\mathcal{A}}_2^\ast \ast \bm{\mathcal{Y}}_2)$, regenerate the \\[-2pt]
\ \ \ random tensor $\bm{\mathcal{A}}_1$, and restart the t-BRP;\\[-2pt]
\ \ \ $\textbf{else}$ continue;\\[-2pt]
\ \ \ $\bm{\mathcal{L}}^{t} =  \bm{\mathcal{Y}}_1 \ast (\bm{\mathcal{A}}_2^{\ast} \ast \bm{\mathcal{Y}}_1 )^{-1} \ast \bm{\mathcal{Y}}_2^\ast$;\\[-2pt]
\ \ \ $\bm{\mathcal{S}}^{t} = \mathcal{H}_{k}(\bm{\mathcal{X}} - \bm{\mathcal{L}}^{t})$;\\[-2pt]
\textbf{End while}\\[-2pt]
  \bottomrule
 \end{tabular}
 \vspace{-0.3cm}
\end{table}
\vspace{-0.1cm}
\section{EXPERIMENTS}
\vspace{-0.15cm}
\label{sec:exp}
\noindent To illustrate the effectiveness of the proposed method, experiments are conducted on the synthetic and the real
data. The compared methods consist of LRTA \cite{LRTA}, BM4D \cite{BM4D}, LRMR \cite{LRMR}, WSNLRMA \cite{WSNLRMA}, and LRTR \cite{LRTR}. The parameters of the compared methods are optimally assigned or selected as suggested in the reference papers.
\vspace{-0.4cm}
\subsection{Synthetic data}
\vspace{-0.2cm}
The HSIs of Washington DC Mall\footnote{http://lesun.weebly.com/hyperspectral-data-set.html\label{web}} $(256 \times 256 \times 191)$ and Pavia University\textsuperscript{\ref{web}}\,$(610 \times 340 \times 103)$ are tested in the simulated experiments.\,The clean datasets are normalized to [0, 1] band-wisely.\,Two noisy datasets are generated as follows.

\begin{figure}[t]
\footnotesize
\setlength{\tabcolsep}{0.99pt}
\begin{center}
\begin{tabular}{cccc}
\includegraphics[width=0.115\textwidth]{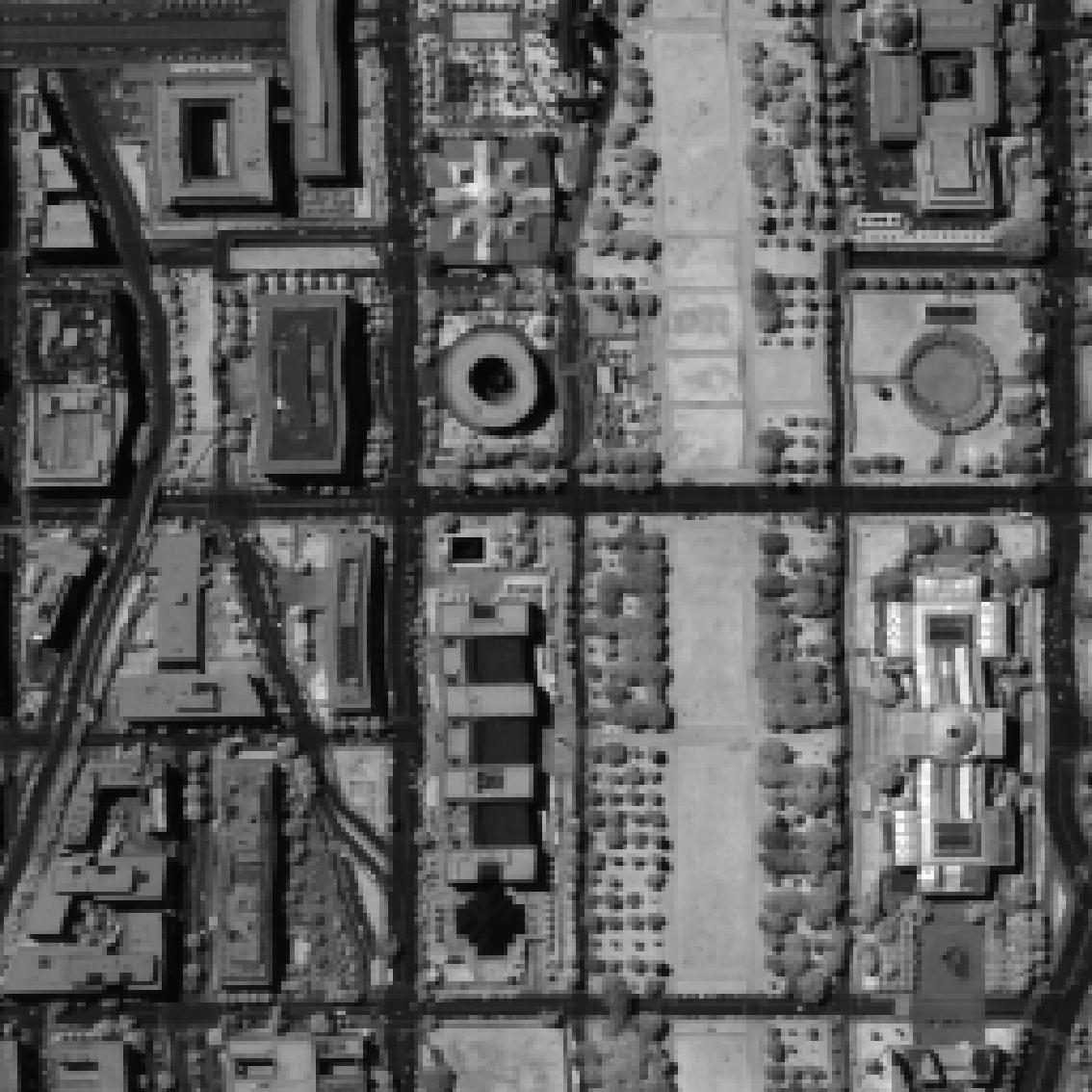}&
\includegraphics[width=0.115\textwidth]{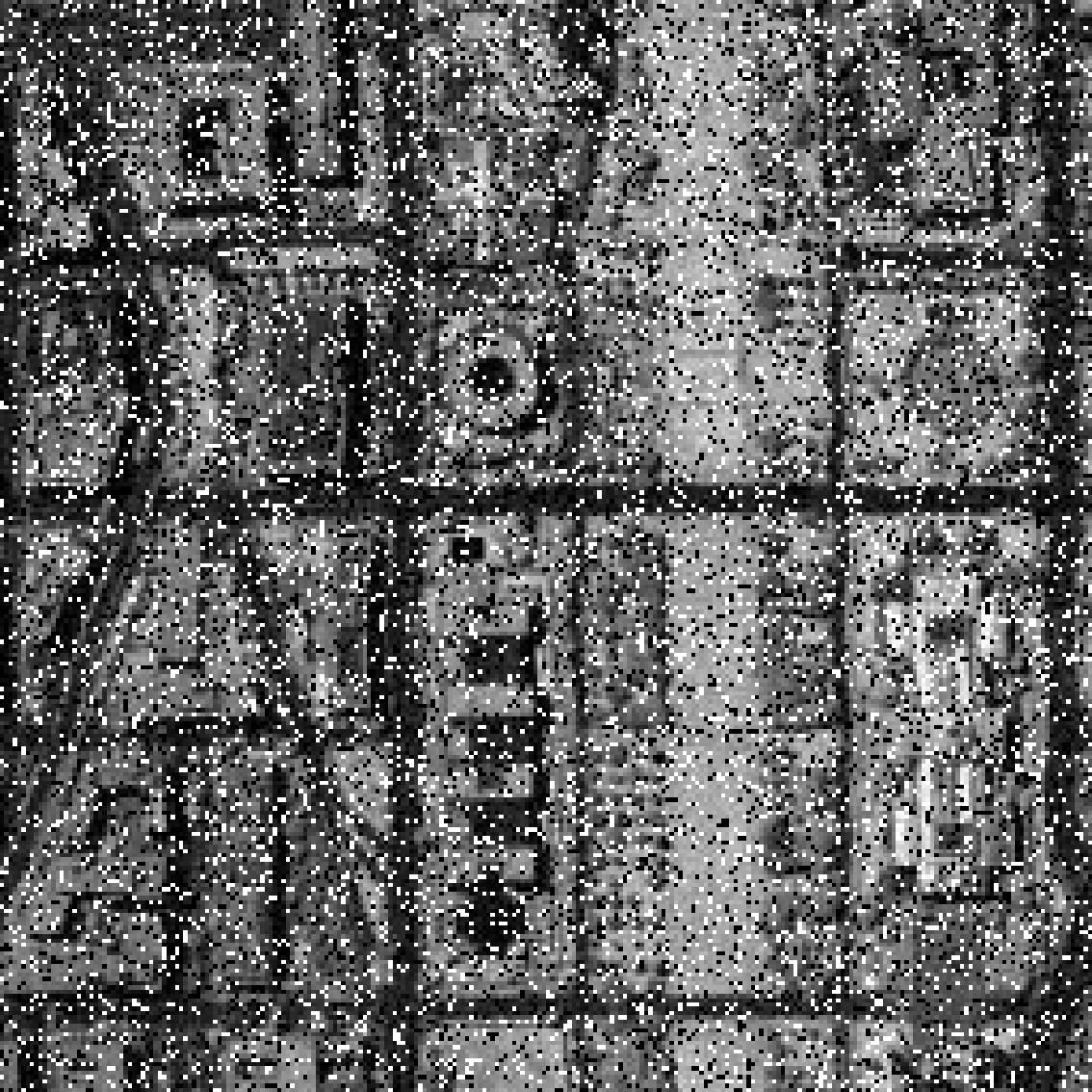}&
\includegraphics[width=0.115\textwidth]{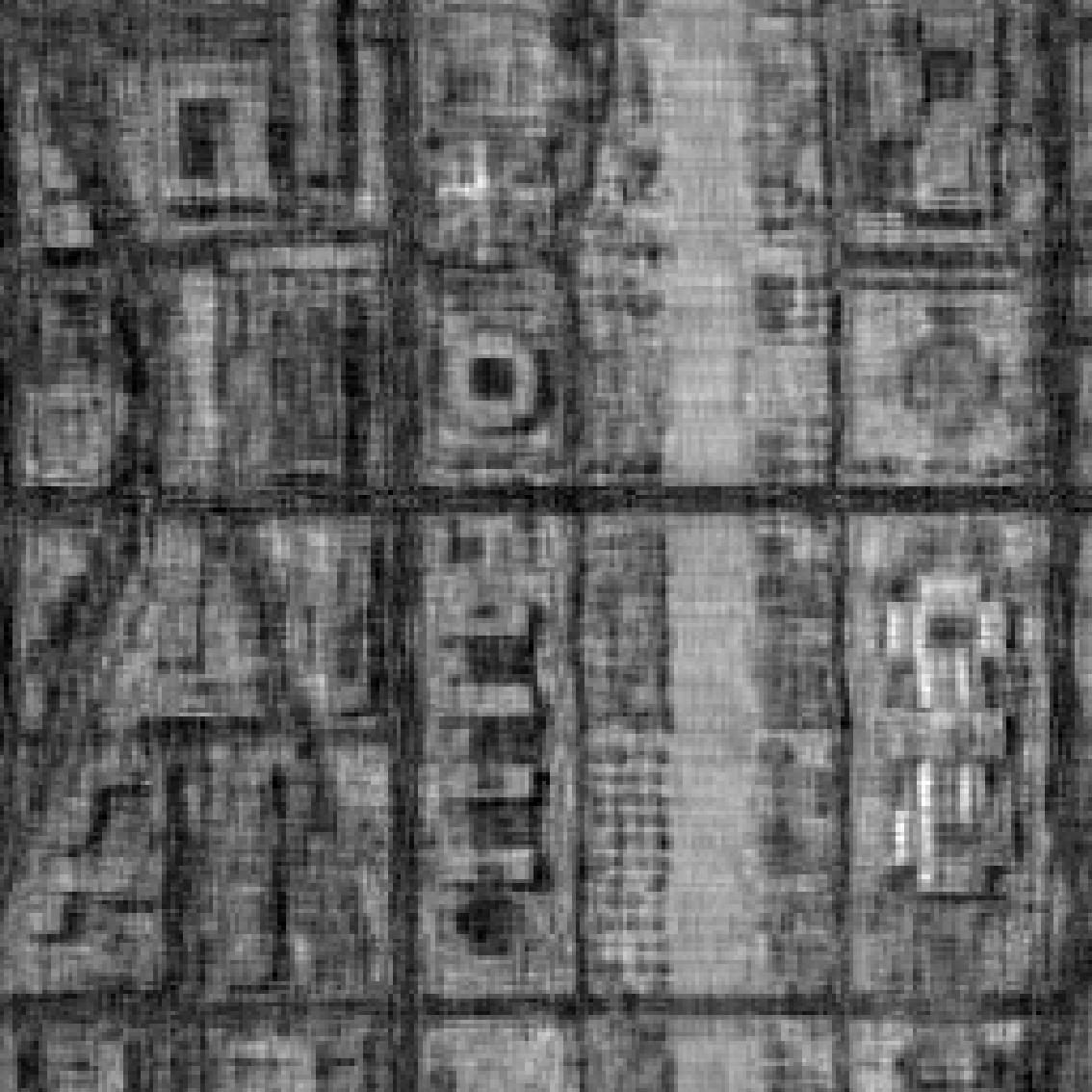}&
\includegraphics[width=0.115\textwidth]{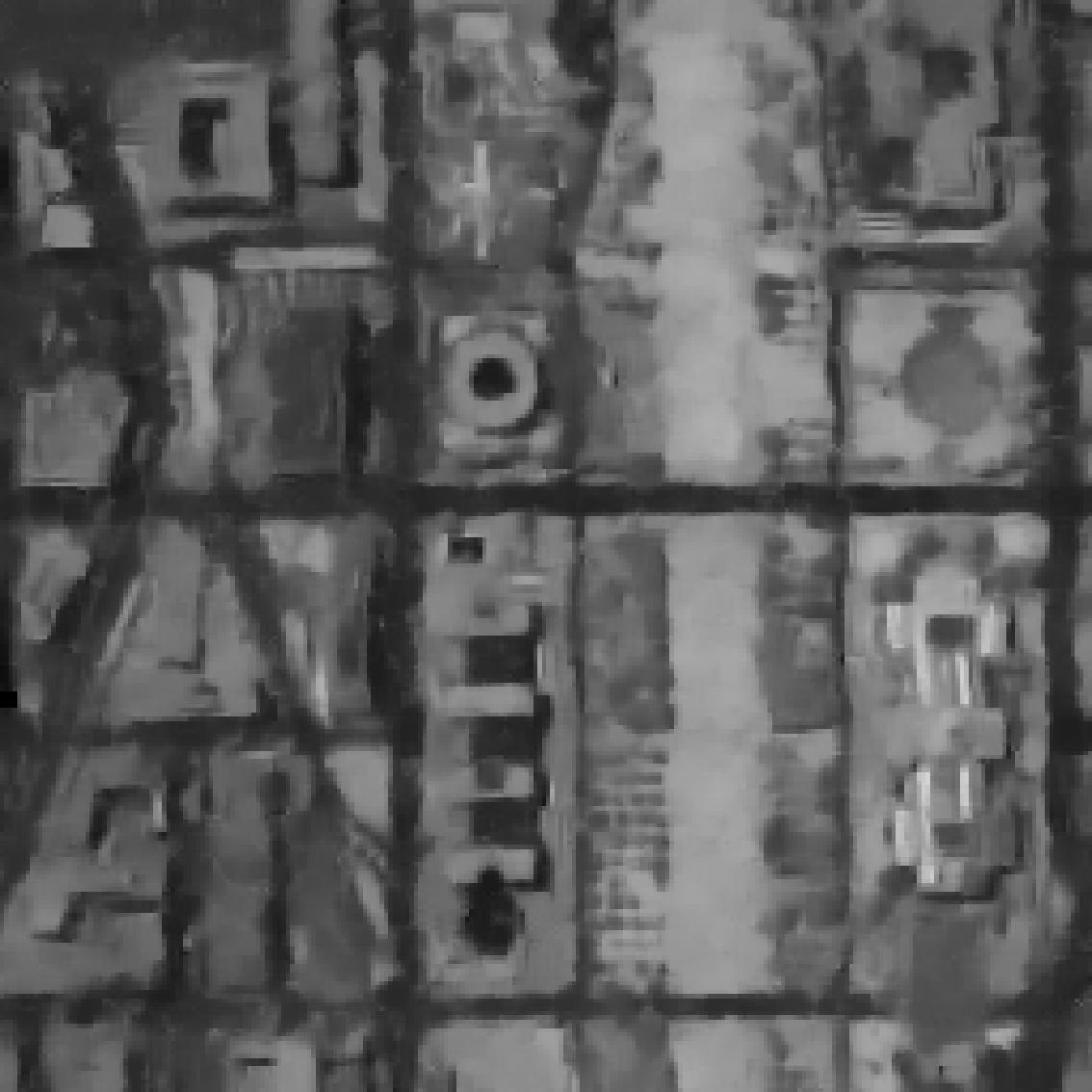}\\
Clean image &  Noisy image &  LRTA &  BM4D \\[0.1cm]

\includegraphics[width=0.115\textwidth]{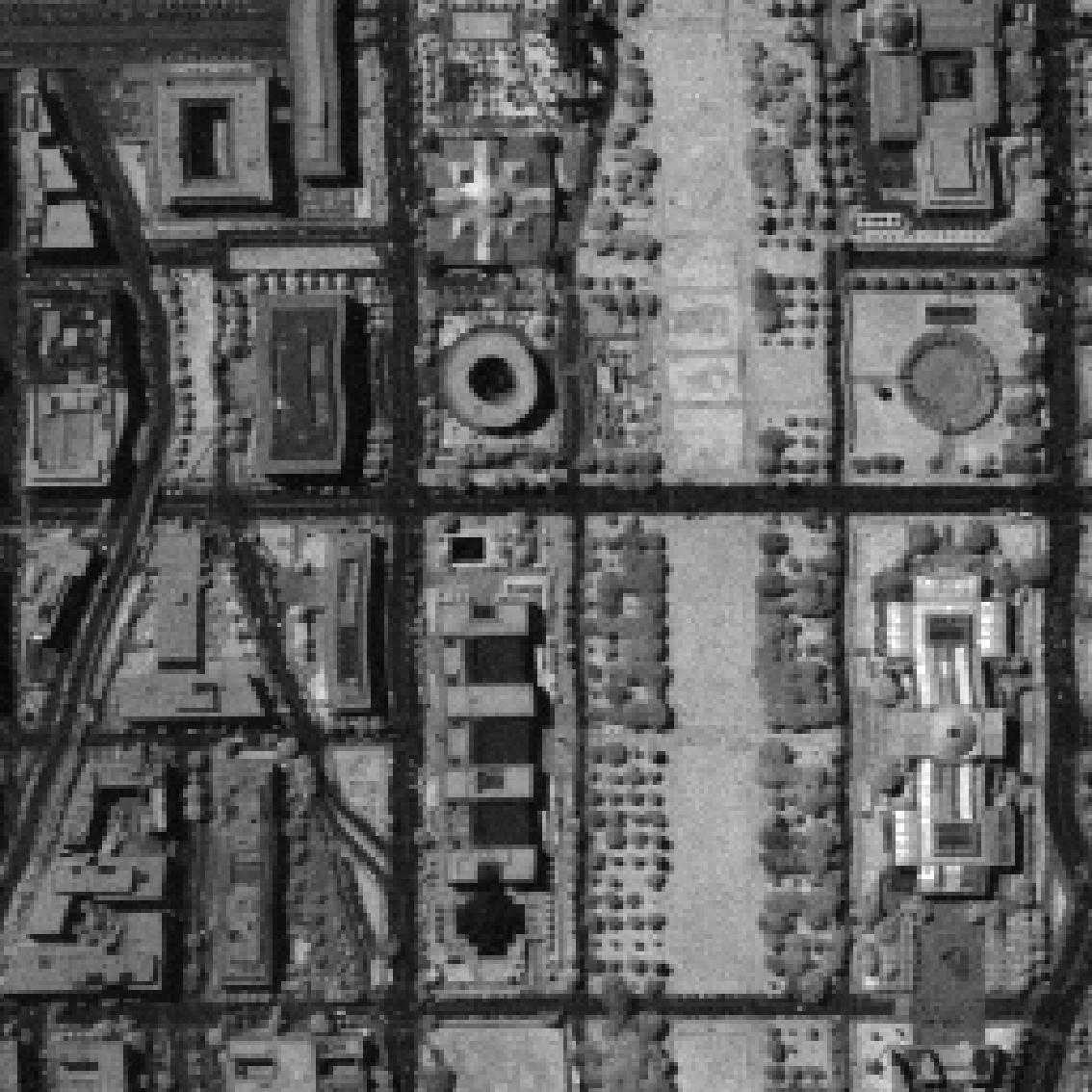}&
\includegraphics[width=0.115\textwidth]{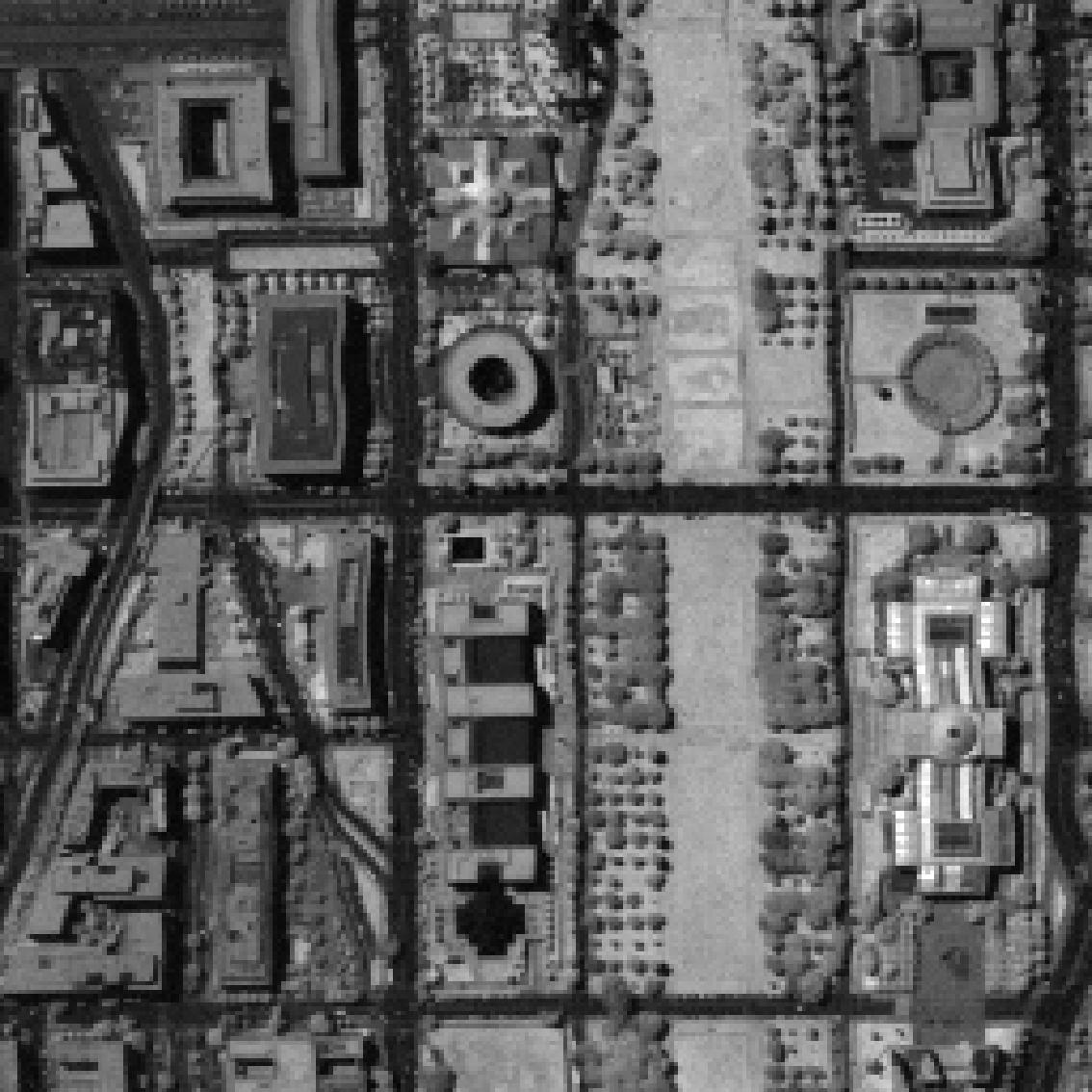}&
\includegraphics[width=0.115\textwidth]{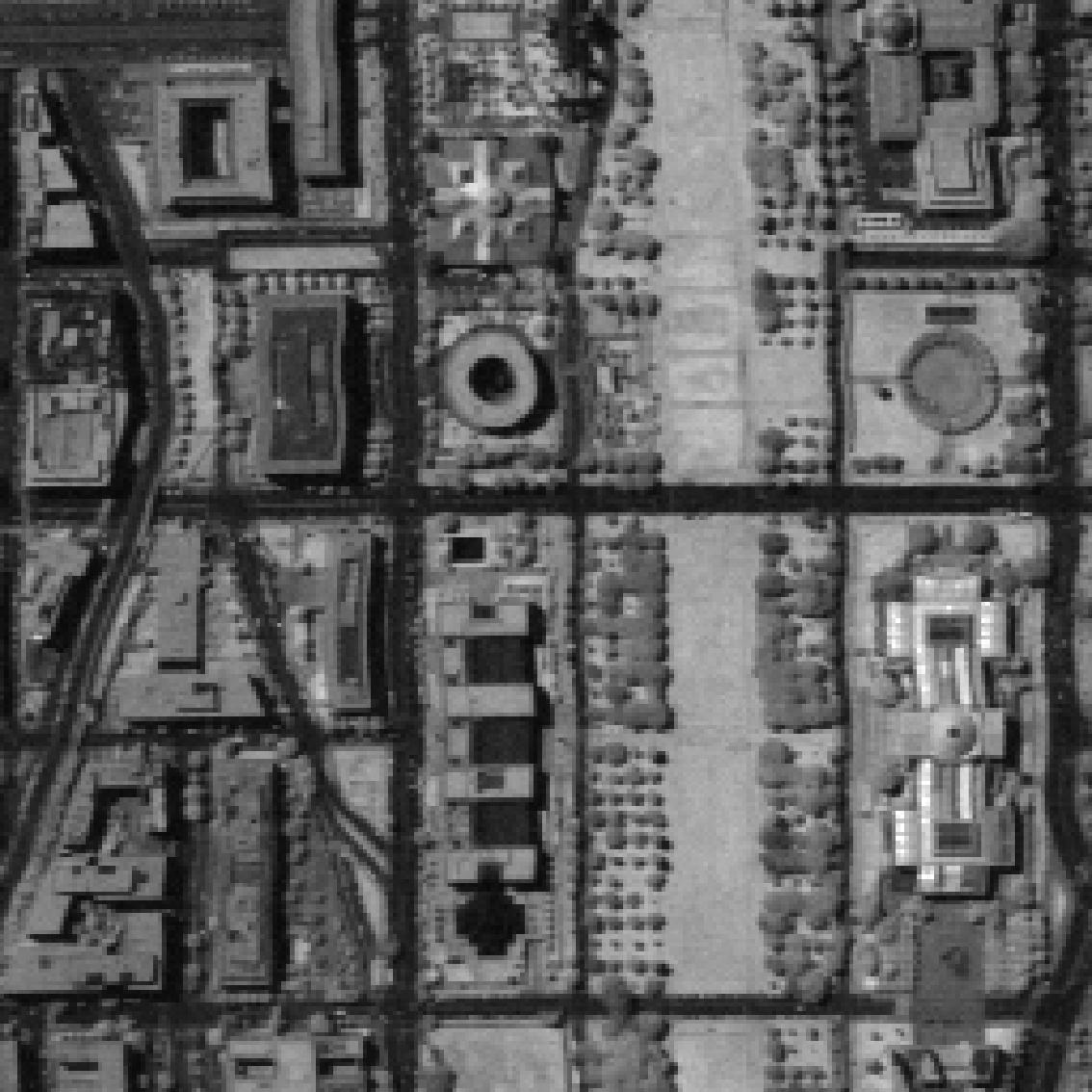}&
\includegraphics[width=0.115\textwidth]{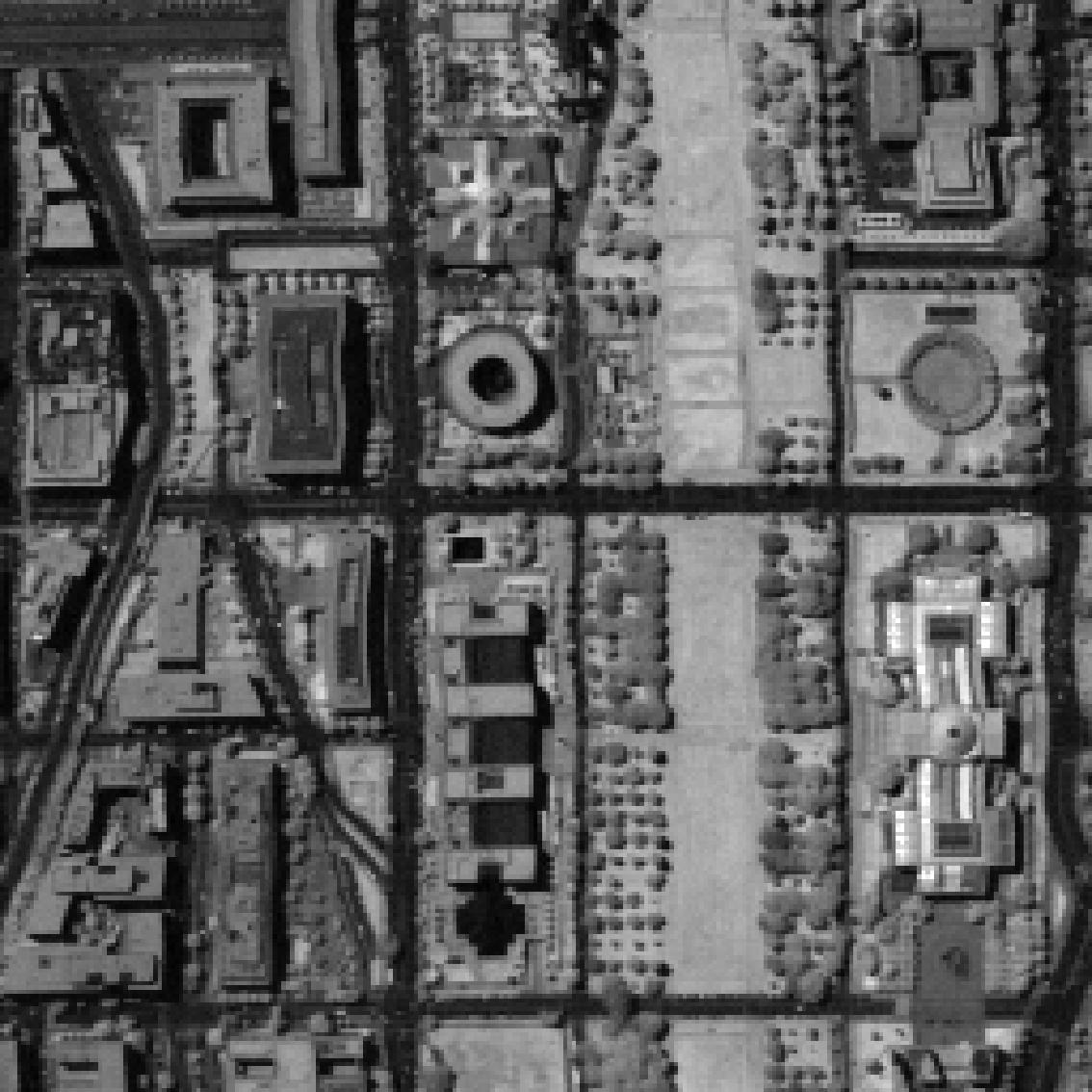}\\
LRMR & WSNLRMA & LRTR & CLTRTR
  \end{tabular}
  \vspace{-0.2cm}
  \caption{Denoising results for band 79 of Washington DC Mall.}
  \label{MSIfig1}
  \end{center}\vspace{-0.55cm}
\end{figure}

\begin{figure}[t]
\footnotesize
\setlength{\tabcolsep}{0.99pt}
\begin{center}
\begin{tabular}{cccc}
\includegraphics[width=0.115\textwidth]{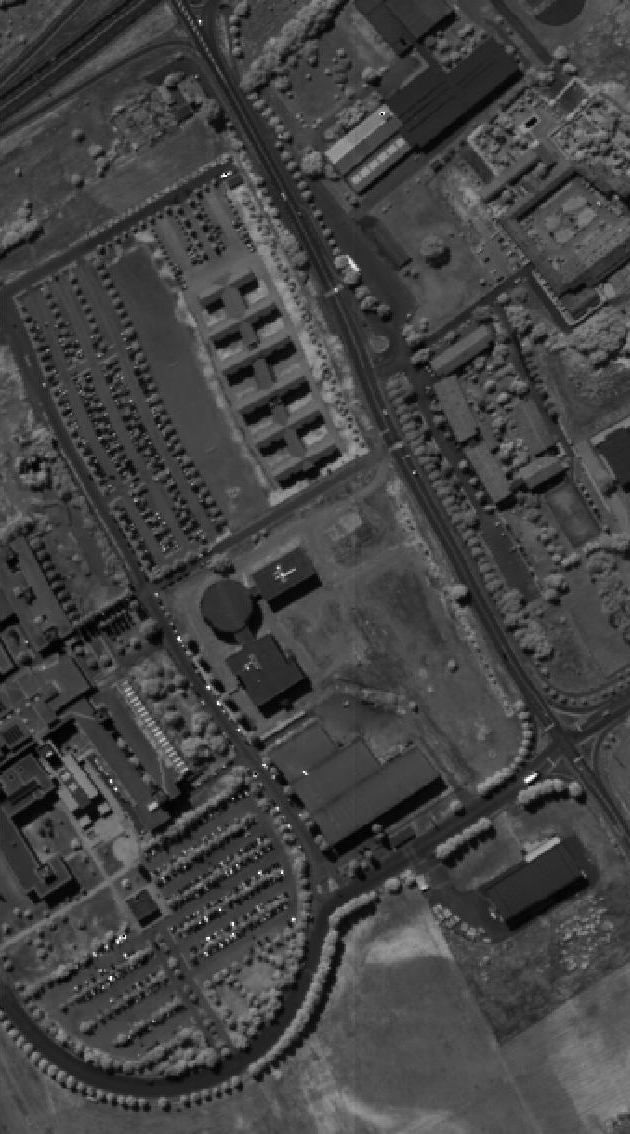}&
\includegraphics[width=0.115\textwidth]{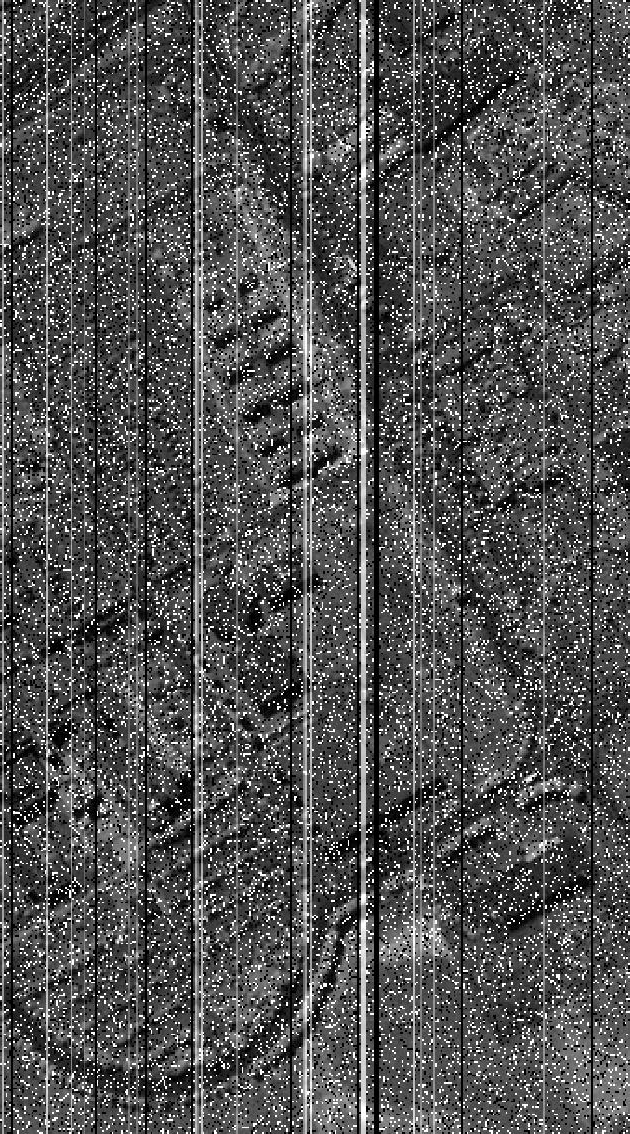}&
\includegraphics[width=0.115\textwidth]{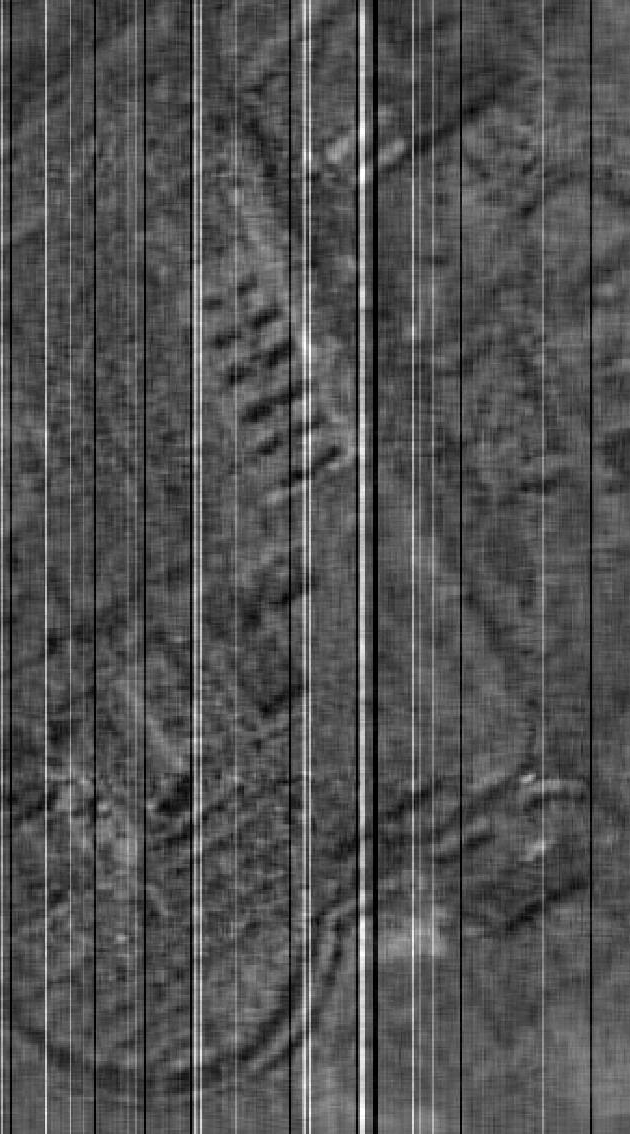}&
\includegraphics[width=0.115\textwidth]{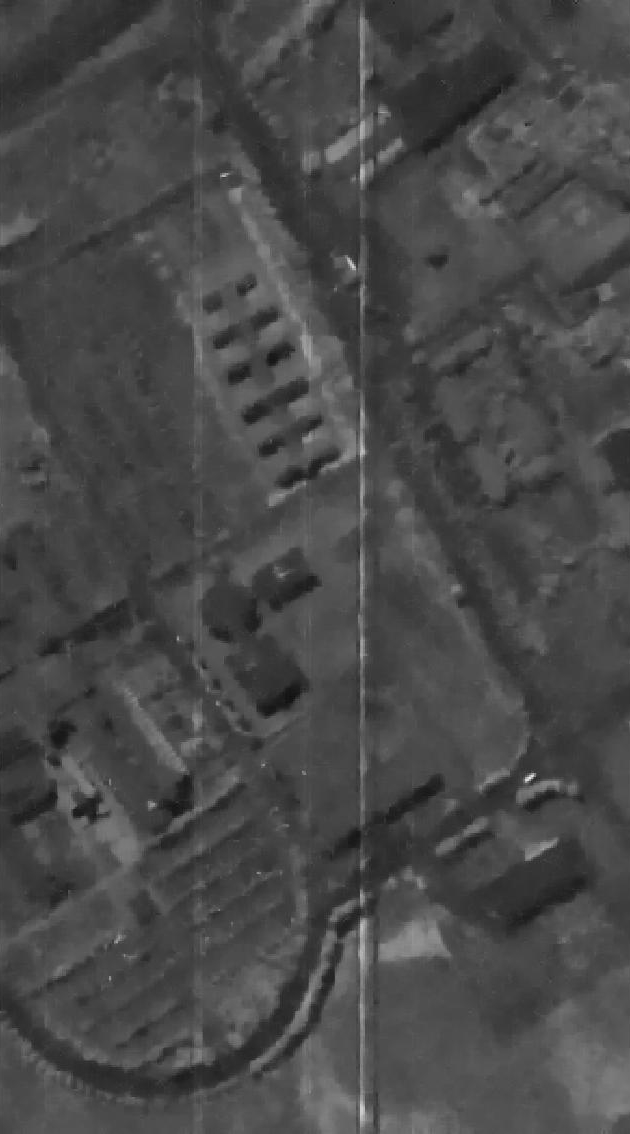}\\
Clean image &  Noisy image &  LRTA &  BM4D \\[0.1cm]

\includegraphics[width=0.115\textwidth]{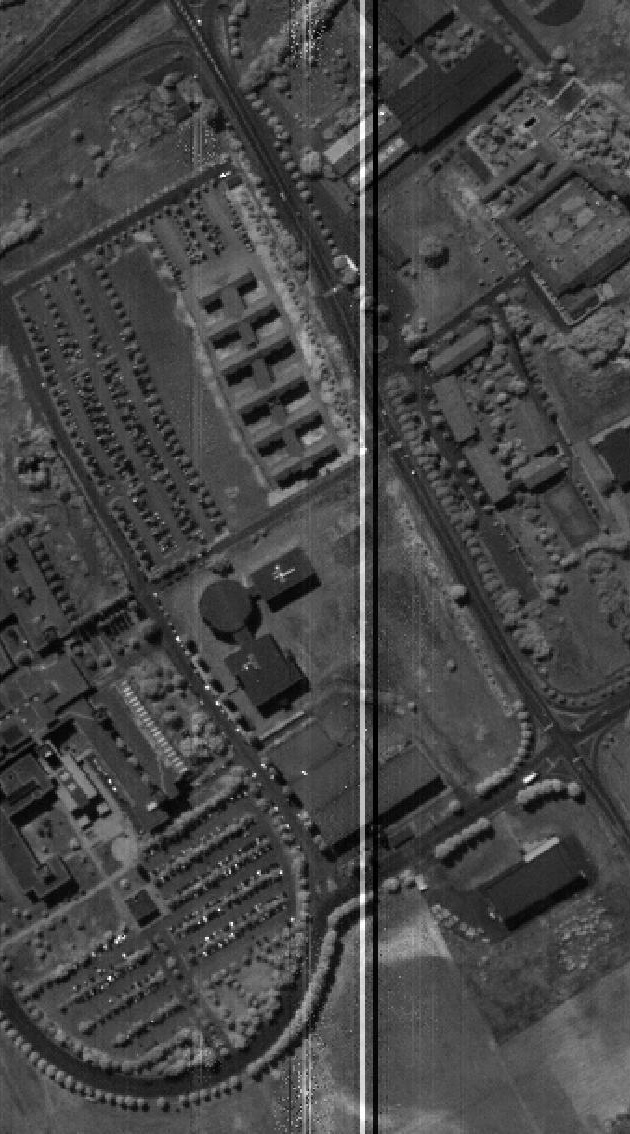}&
\includegraphics[width=0.115\textwidth]{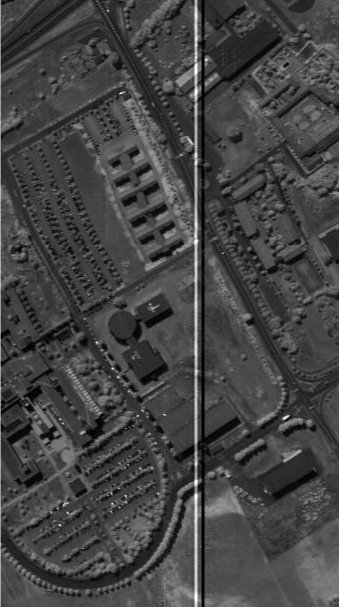}&
\includegraphics[width=0.115\textwidth]{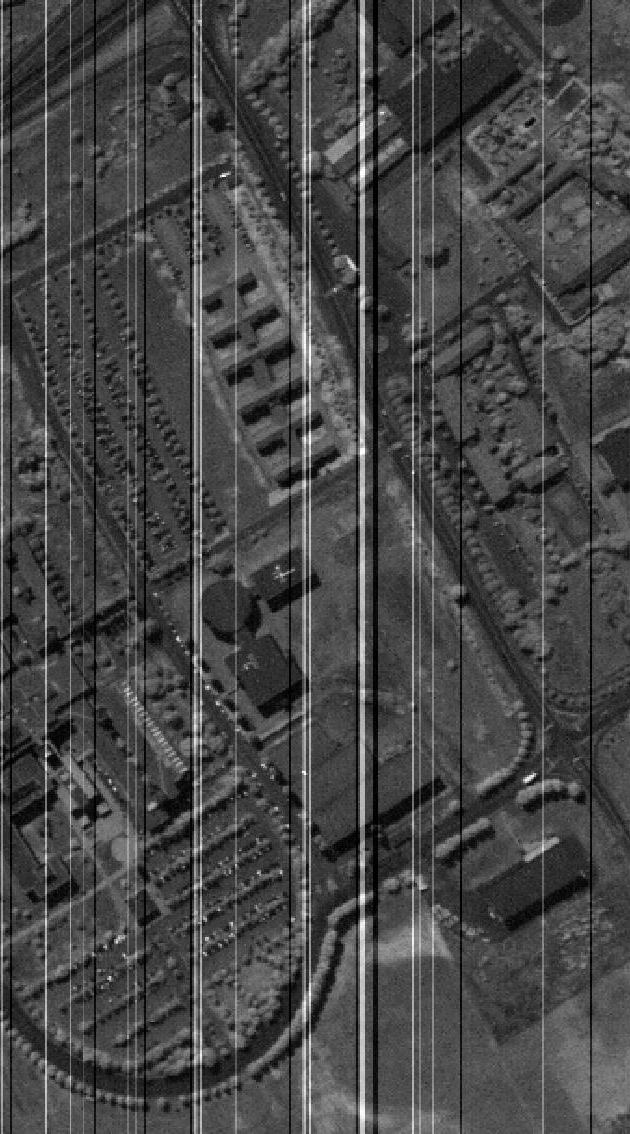}&
\includegraphics[width=0.115\textwidth]{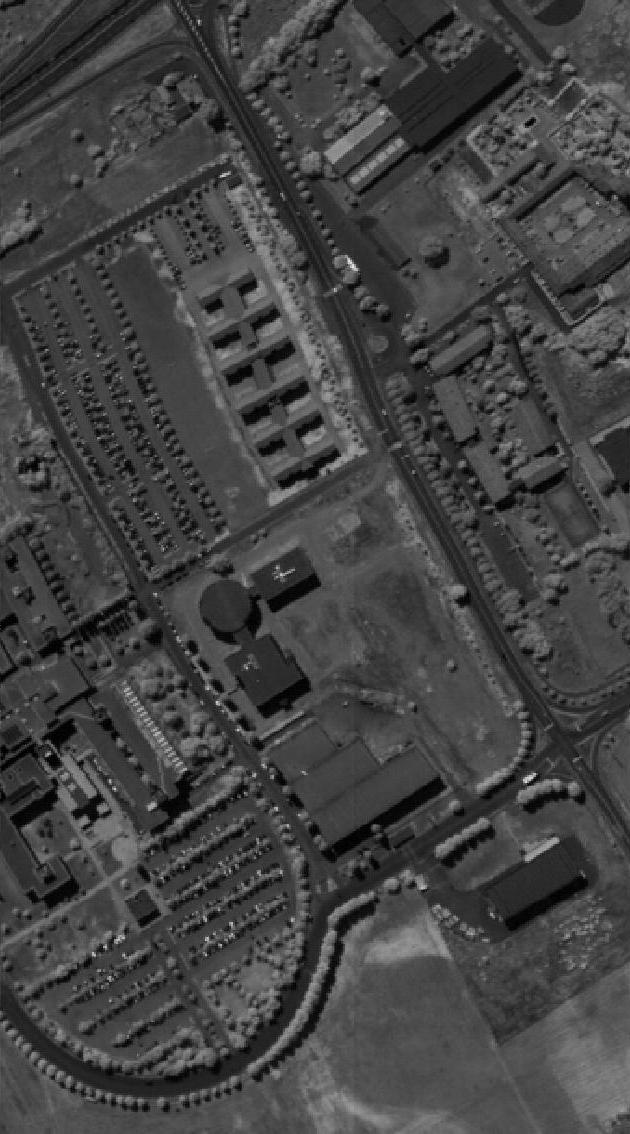}\\
LRMR & WSNLRMA & LRTR & CLTRTR
  \end{tabular}
  \vspace{-0.2cm}
  \caption{Denoising results for band 79 of Pavia University.}
  \label{MSIfig2}
  \end{center} \vspace{-0.9cm}
\end{figure}

\textbf{Case 1:} zero-mean Gaussian noise with variance $\sigma = 0.04$
and the sparse noise (salt and pepper) which affects 20\% pixels are added to all bands of Washington DC Mall.

\textbf{Case 2:} zero-mean Gaussian noise with variance $\sigma = 0.02$
and the sparse noise which affects 20\% pixels are added to all
bands of Pavia University. Then stripes and
deadlines are simultaneously added to 10 arbitrarily selected
bands from this dataset with the width from one line to three
lines.

\begin{table*}[!t]
%\tiny
%\small
\footnotesize
%\small
\setlength{\tabcolsep}{10pt}
\renewcommand\arraystretch{1}\vspace{-0.22cm}
\caption{The performance comparison of six competing methods.}\vspace{-0.6cm}
\begin{center}
%\begin{tabular}{p{0.5cm}p{0.45cm}|p{0.2cm}p{0.1cm}p{0.45cm}|p{0.2cm}p{0.1cm}p{0.45cm}|p{0.2cm}p{0.1cm}p{0.45cm}}
\begin{tabular}{cccccccccc}
 \toprule
  & Data & Index & Noisy & LRTA & BM4D & LRMR & WSNLRMA & LRTR & CLTRTR\\
  \hline
  \multirow{4}{*}{Case 1} & \multirow{4}{*}{Washington DC Mall}  & MPSNR & 11.31&20.42 &22.78 &34.86&35.15&35.79&\textbf{37.69}\\
&& MSSIM & 0.117 &0.478 &0.518&0.947 &0.959&0.955&\textbf{0.969}\\
&& SAM & 47.52&17.66&13.78&	6.18&5.80&4.81&	\textbf{4.09}\\
&& Time (s) & -&\emph{37.44} &102.36&194.11&3871.23&\underline{181.75}&\underline{103.70}\\
  \hline

\multirow{4}{*}{Case 2}   &  \multirow{4}{*}{Pavia University}   & MPSNR & 11.31&20.78&24.21 &33.51 &35.71&34.91&\textbf{39.18}\\
&& MSSIM & 0.078 &0.450 &0.520&0.912 &0.924&0.903&\textbf{0.969}\\
&& SAM & 47.49&15.58 &10.18 &6.03 &5.71&7.98&\textbf{2.94}\\
&& Time (s) & -&\emph{51.39}&163.25&475.98&13524.52&\underline{268.74}&	\underline{173.26}\\
\toprule
\end{tabular}
\end{center}\vspace{-0.4cm}
\label{MSItab}
\end{table*}

\begin{figure*}[!t]
\vspace{-0.1cm}
\footnotesize
\setlength{\tabcolsep}{0.97pt}
\begin{center}
\begin{tabular}{ccccccc}
\includegraphics[width=0.137\textwidth]{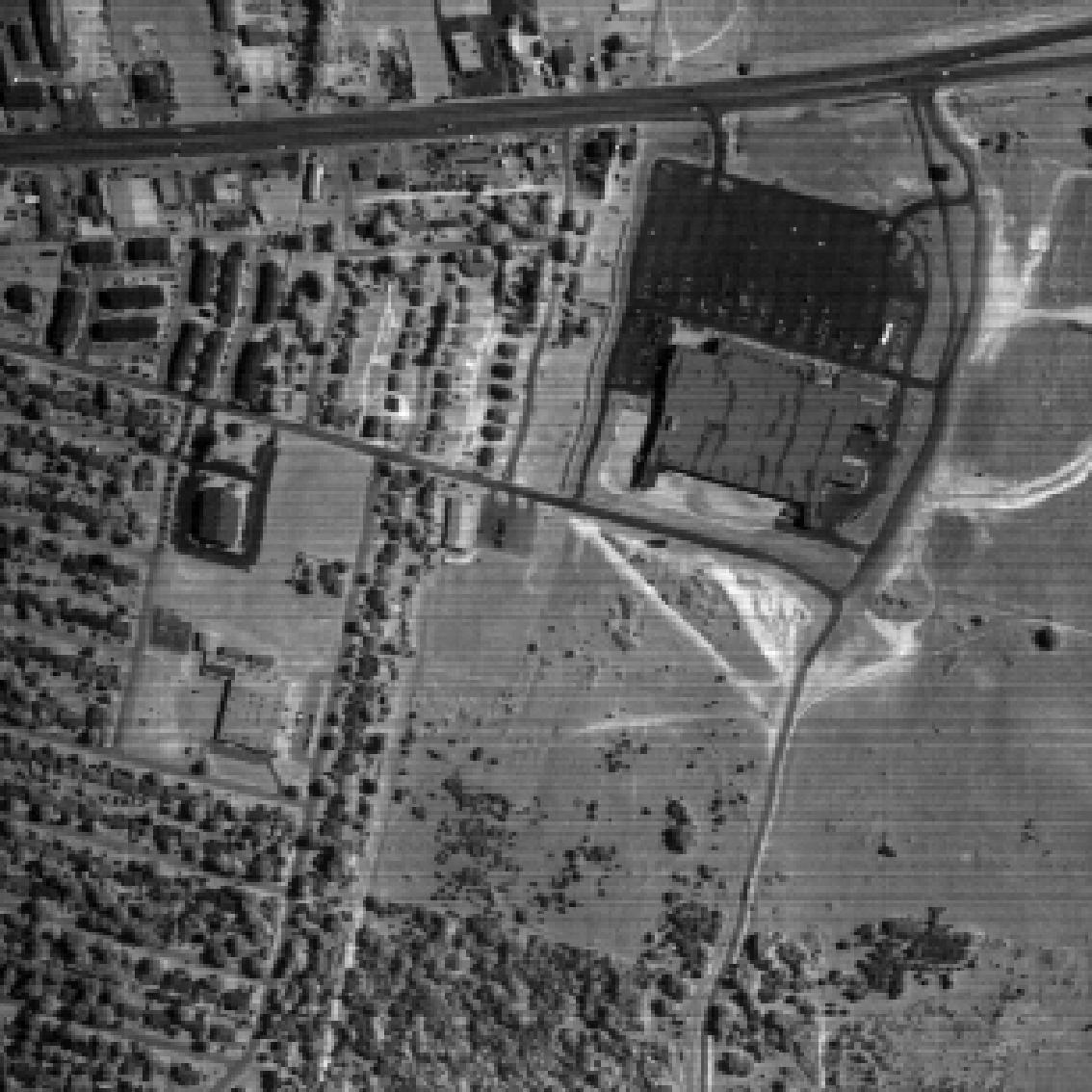}&
\includegraphics[width=0.137\textwidth]{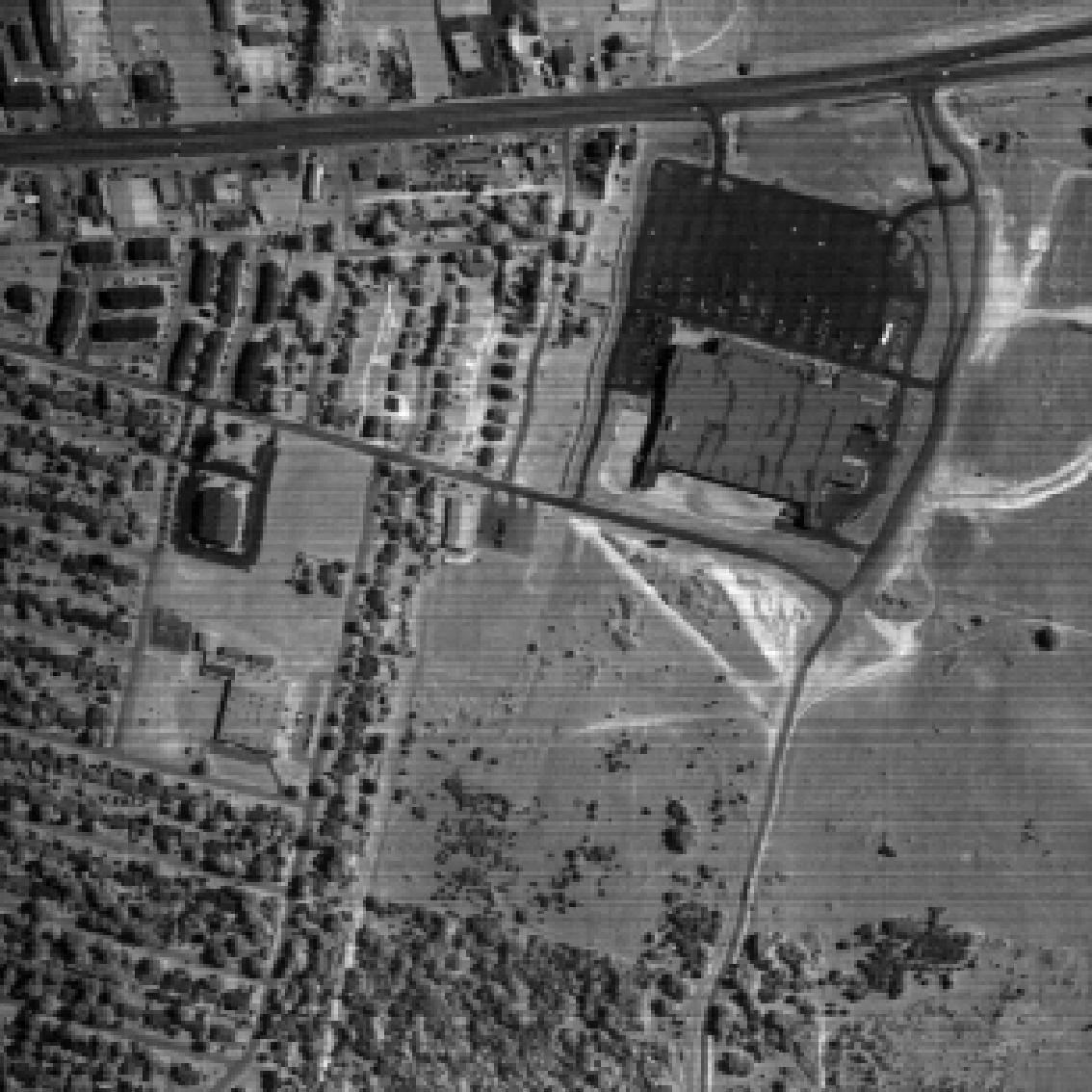}&
\includegraphics[width=0.137\textwidth]{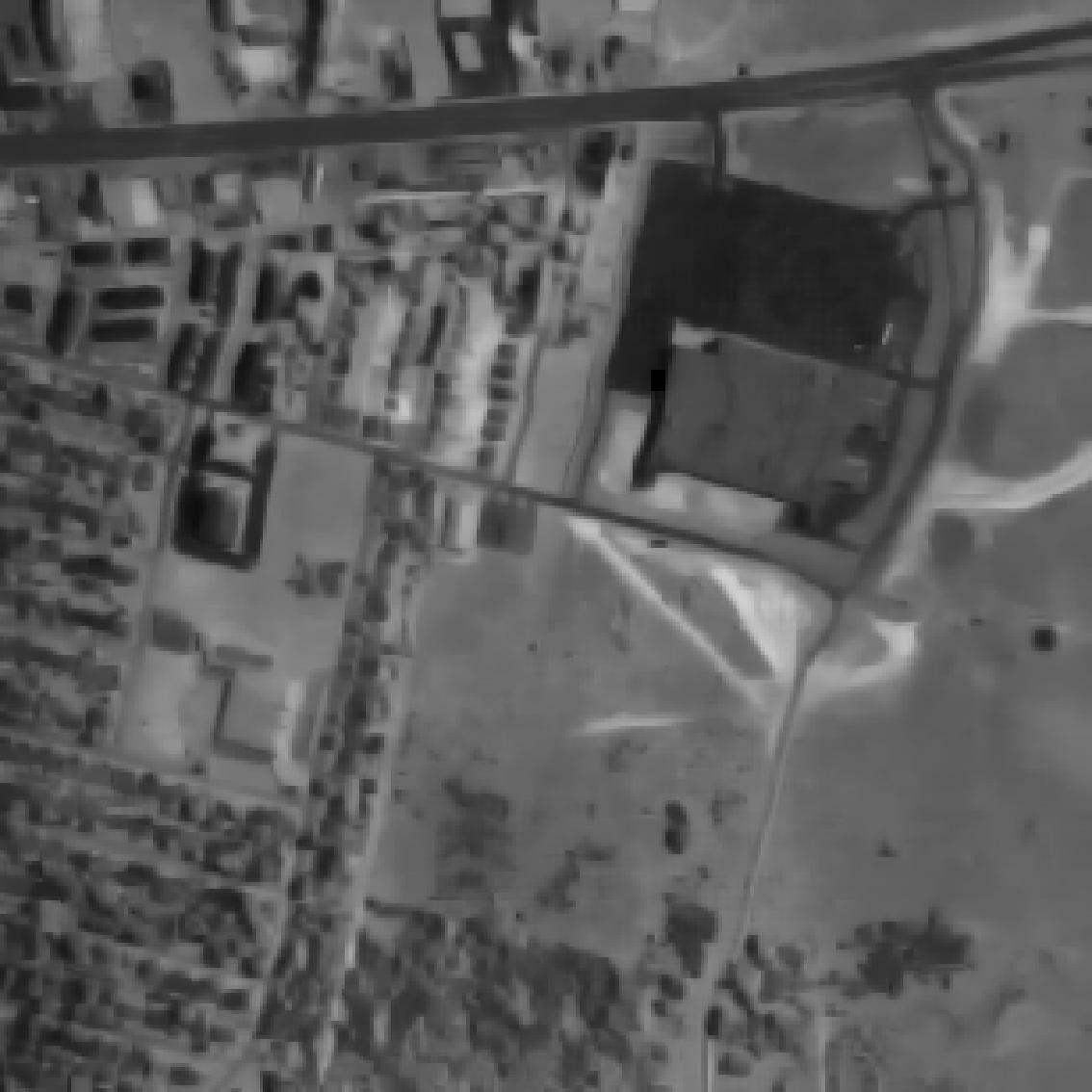}&
\includegraphics[width=0.137\textwidth]{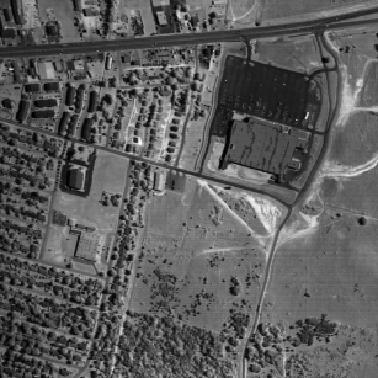}&
\includegraphics[width=0.137\textwidth]{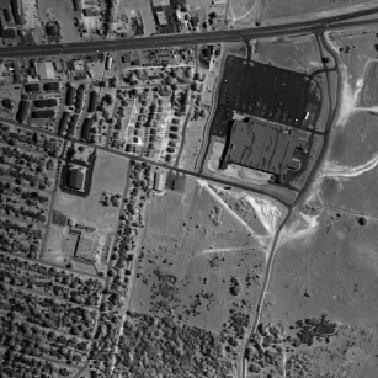}&
\includegraphics[width=0.137\textwidth]{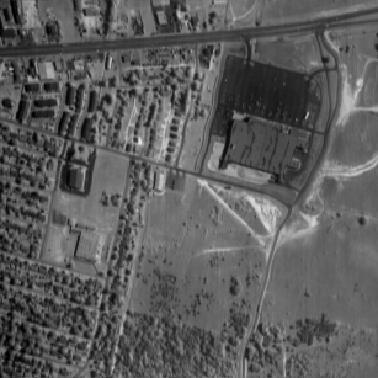}&
\includegraphics[width=0.137\textwidth]{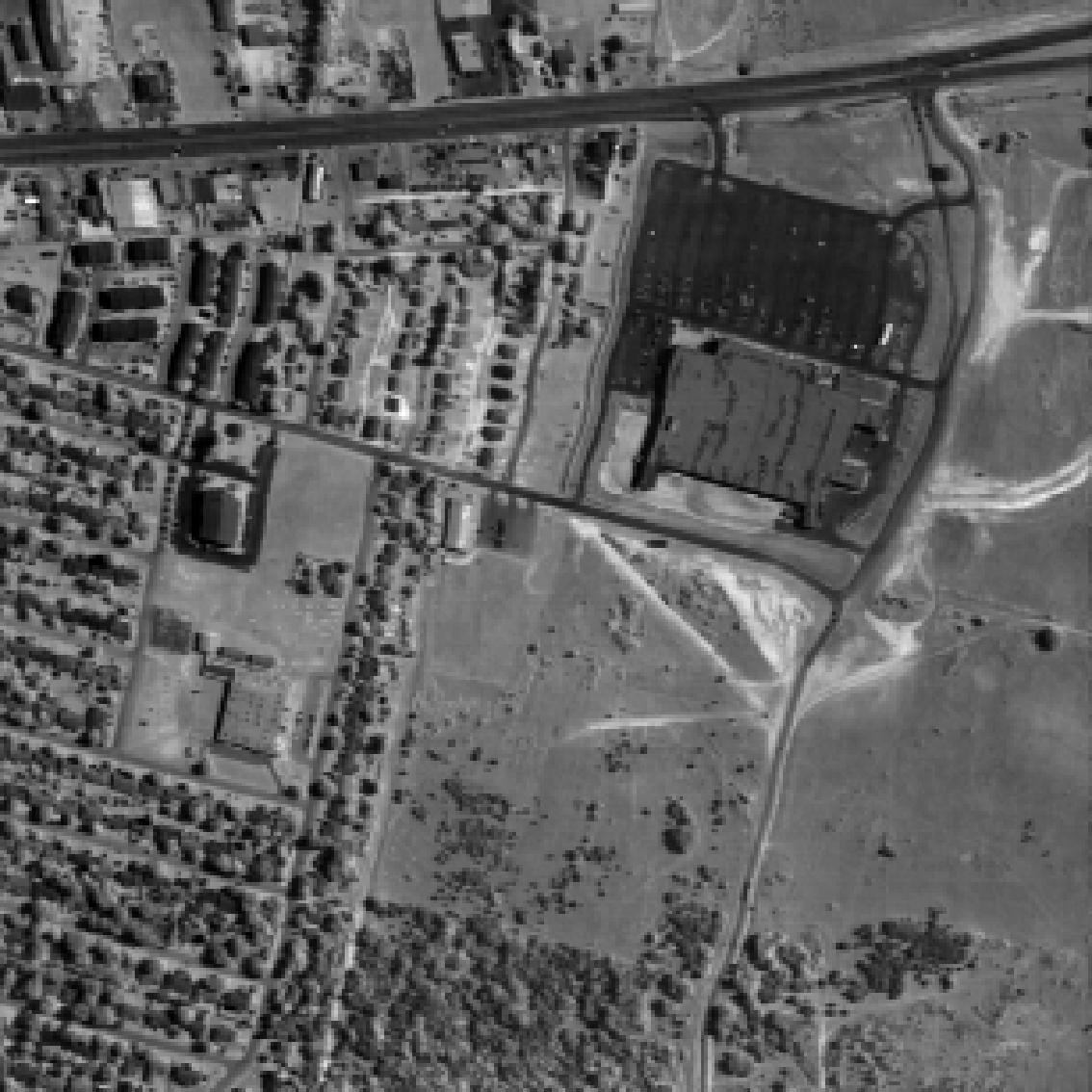}\\
Noisy image &  LRTA &  BM4D & LRMR & WSNLRMA & LRTR & CLTRTR
  \end{tabular}
  \vspace{-0.2cm}
  \caption{Denoising results for band 103 of HYDICE urban.}
  \label{MSIfig3}
  \end{center}\vspace{-0.8cm}
\end{figure*}
For the proposed method, the parameters $r$ and $k$ are set to be $5$ and $2\times 10^6$, respectively. The peak signal-to-noise ratio (PSNR), the structural similarity index (SSIM), and the
spectral angle mapping (SAM) are employed for performance evaluation. Table \ref{MSItab} lists the quantitative
comparisons. As observed, the proposed method outperforms the compared ones with respect to PSNR, SSIM, and SAM values. For the running time (in seconds), the proposed method is the third fastest, but considering the methods with results PSNR higher than 30dB, i.e., LRMR, WSNLRMA, and LRTR, the proposed method is the fastest.

Denoising results are illustrated in
Fig.\ref{MSIfig1} and Fig.\ref{MSIfig2}. It can be observed that the proposed method achieves the best visual results among all compared methods. Specially, in case 1, all compared methods perform well in removing mixed noise, except that LRTA and BM4D cause details missing. In case 2, the results obtained by the compared methods contain a small number of stripes, where the proposed method removes almost the mixed noise.

\vspace{-0.3cm}
\subsection{Real data}
\vspace{-0.1cm}
The imagery of HYDICE urban data\footnote{http://www.tec.army.mil/hypercube} is used in the experiment. The size of original dataset is $304 \times 304 \times 210$. The bands 104-108, 139-151, and 208-210 are removed due to serious pollution by the atmosphere and water absorption. The denoising results for band
103 are illustrated in Fig.\ref{MSIfig3}. It can be seen that our method removes almost all stripes and finely preserves the intrinsic structure. LRMR and WSNLRMA achieve great performance but destroy the partial-spatial structure. LRTR effectively removes stripes, but its result contains evident loss of details.
\vspace{-0.2cm}
\section{CONCLUSION}
\vspace{-0.2cm}
\label{sec:concl}
\noindent In this paper, we introduced a novel constrained low-tubal-rank tensor recovery model to remove mixed noise in HSIs. Then we developed a t-BRP algorithm for efficiently solving the proposed model. The synthetic and real data experiments demonstrate that our method achieved excellent performance on
HSIs mixed noise removal and preserved the structure. Moreover, the running time comparison indicated the high efficiency of our algorithm.

\footnotesize
\renewcommand{\refname}{\small \center \vspace{-0.9cm}REFERENCES\vspace{-0.1cm}}
\setlength{\bibsep}{0.05ex}
\bibliographystyle{IEEEtran}
\bibliography{refs}
\end{document}